\documentclass[letterpaper, 10 pt, conference]{ieeeconf}  

\IEEEoverridecommandlockouts                              

\usepackage{hyperref}									
\usepackage{graphicx} 									
\usepackage{times}										 
\usepackage{amsmath} 									
\usepackage{amssymb} 									 
\usepackage{grffile}
\usepackage{caption}
\usepackage[font={scriptsize}]{caption}
\usepackage{subcaption}
\usepackage{tikz}
\usetikzlibrary{positioning,shapes,shadows,arrows}
\usepackage{amsmath,tikz}
\usepackage[export]{adjustbox}
\usepackage{footnote}
\usepackage{breqn}

\hypersetup{linkbordercolor=green}

\usepackage[font=small,skip=0pt]{caption}

\tikzstyle{decision} = [diamond, draw, aspect =1.8, fill=pink!15,
text width=6em, text badly centered, node distance=4.5cm, inner sep=2pt]
\tikzstyle{block} = [rectangle, rounded corners, draw = blue, thick, fill=blue!16,
text width=9em, text centered, minimum height=3em]
\tikzstyle{line} = [draw, very thick, color=black!60, -latex']
\tikzstyle{cloud} = [draw, ellipse,fill=purple!20, node distance=3.0cm,
minimum height=1em]

\title{\LARGE \bf
	Real-Time, Soft Robotic Patient Positioning System for Maskless Head-and-Neck Cancer Radiotherapy*
}

\author{Olalekan P. Ogunmolu$^{1}$, Xuejun Gu$^{2}$, Steve Jiang$^{2}$, and Nicholas R. Gans$^{1}$  
	\thanks{*This work was supported by the Radiation Oncology Department, UT Southwestern, Dallas, Texas, USA}
	\thanks{$^{1}$Olalekan P. Ogunmolu and Nicholas R. Gans are with the Department of Electrical Engineering,
		University of Texas at Dallas, Richardson, TX 75080, USA
		{\tt\small \{olalekan.ogunmolu, ngans\}@utdallas.edu}}%
	\thanks{$^{2}$Xuejun Gu and Steve Jiang are with the Department of Radiation Oncology,  
		University of Texas Southwestern Medical Center, Dallas TX 75390, USA
		{\tt\small \{Xuejun.Gu, Steve.Jiang\}@utsouthwestern.edu}}%
}

\begin{document}
	
	\long\def\/*#1*/{}								
	
	\graphicspath{ {Charts/} }
	
	\maketitle
	\thispagestyle{empty}
	\pagestyle{empty}

	\begin{abstract}
		
		We present an initial examination of a novel approach toward accurately positioning a patient during head and neck intensity modulated radiotherapy (IMRT). Position-based visual-servoing of a radio-transparent soft robot is used to control the flexion/extension cranial  motion of a manikin head. A Kinect RGB-D camera is used to measure head position and the error between the sensed and desired position is used to control a pneumatic system which regulates pressure within an inflatable air bladder (IAB). Results show that the system is capable of controlling head motion to within 2mm with respect to a reference trajectory. This establishes proof-of-concept that using multiple IABs and actuators can improve cancer treatment.
		

	\end{abstract}

	\section{Introduction}
	This paper presents a systematic initial examination of an image-guided soft robot patient positioning system for use in head and neck (H\&N) cancer radiotherapy (RT). H\&N cancers are among the most fatal of major cancers. In 2014, 1,665,540 new patients developed pharynx and oral cavity cancers which led to 585,720 deaths in the United States \cite{c1}. Treating these cancers often involve intensity modulated radiotherapy (IMRT) where a patient lies on a 6-DOF movable treatment couch and lasers or image-guiding systems are used to ensure the patient is in the proper position. A linear accelerator (LINAC) is used to accelerate electrons in a wave guide to enable collision of electrons with a heavy metal target. High-energy x-rays produced from the collisions are shaped by multileaf collimators as they exit the gantry of the machine to conform to the shape of the patient's tumor. The beam that emerges can be directed to a tumor from any angle by rotating the gantry and moving the couch.
	
	IMRT requires accurate patient positioning while high potent dose radiation is delivered to tumor while sparing critical organs nearby. An examination of dosimetric effects on patient displacement and collimator and gantry angle misalignment during IMRT showed high sensitivity to small perturbations: a 3-mm error in anterior-posterior direction caused 38\% decrease in minimum target dose or 41\% increase in the maximum spinal cord dose \cite{c2}. Treatment discomfort and severe pain often results from long hours of minimally invasive surgery where the skull is fixed with pins for head immobilization during stereo-tactic radiosurgery (SRS). In addition, conventional linear accelerators (LINACs) used at most cancer centers are insufficient for the high geometric accuracy and precision required of SRS for isocenter localization \cite{c7}. 
	\/*
	Stereo-tactic Radiotherapy (SRT) is a cancer treatment method which requires a rigid immobilization system to achieve high geometric accuracy and precision for isocenter localization. Alternative high-accuracy fractionated SRT treatment include invasive SRT head frames which rigidly fix the geometry of intracranial lesion coordinates; and noninvasive SRT systems using masks, non-rigid frames and implanted fiducials to achieve a high-degree of accuracy.*/ 
	
	Image-guided radiotherapy (IGRT) has made progress in improving accuracy while reducing set-up times \cite{c7, c8, c6}. The Robotic Tilt Module (RTM) interfaced with an image-guidance system  \cite{c7, RTM} enables high-precision positional correction by automatically aligning the patient when the image-guidance system detects positional errors. However, the power of IGRT hasn't been fully explored due to the limited degrees of freedom of couch motion. State-of-the-art couches can only correct rigid errors, but not compensate for curvature changes, which often occurs in  neck positioning. Also, patient motions are often ignored during image-guidance procedures, where the focus is on the use of images only before treatment. 	
	\/*
	The Novalis system can detect rotational set-up errors with an average accuracy of $0.09^{\circ}$ (standard deviation, $\sigma$, $0.06^{\circ}$), $0.02^{\circ}$ ($\sigma$, $0.07^{\circ}$) and $0.06^{\circ}$ ($\sigma$, $0.14^{\circ}$) for longitudinal, lateral and vertical rotations. It has an average positioning accuracy of $0.06^{\circ}$ ($\sigma$, $0.04^{\circ}$), $0.08^{\circ}$ ($\sigma$, $0.06^{\circ}$) and $0.08^{\circ}$ ($\sigma$, $0.07^{\circ}$) for longitudinal, lateral and vertical rotations respectively \cite{RTM}. 
	
	Recently, a 6D robotic real-time surface image-guided positioning system was employed in a feasibility study for frameless and maskless cranial stereotactic radiosurgery (SRS) \cite{cervino}. It tested the accuracy of a surface-image guided procedure against an optical guidance platform (OGP) currently used to treat SRS at many institutions. While patient comfort and minimal immobilization at the same accuracy as the OGP was achieved, the procedure emphasized an  inspection based vision system and cooperation of patient to achieve immobilization. We consider this impractical since it assumes all patient cooperation and involve treatment interruptions whenever patient motion  is beyond pre-defined tolerance.
	
	Results from the clinical studies \cite{cervino, c2, c3} serve as motivation for minimizing patient positioning errors during IGRT in order to minimize positioning-related uncertainties in cancer dose treatments, improve tumor control, and reduce toxicity \cite{c4, c5}.
	*/
	
	The overall goal of this work is to address the non-rigid motion compensation during H\&N RT. For an initial investigation, we control the one degree of freedom, raising or lowering of a generic patient's head, lying in a supine position, to a desired height above a table. The system consists of a single inflatable air bladder (IAB), a mannequin head and a neck/torso motion simulator, Kinect Xbox 360 Sensor, two pneumatic valve actuators controlled by custom-built current regulators, and a National Instruments myRIO microcontroller. The Kinect Sensor is mounted directly above the head for displacement measurement. The error between the measured and desired head position, as sensed by the camera, is used in a PI controller nested within a PID feedforward to control the pneumatic actuator valves, thereby regulating air pressure within the IAB and moving the patient's head.
	
	Soft robot systems are deformable polymer enclosures with fluid-filled chambers that enable manipulation and locomotion tasks  by a proportional control of the amount of fluid in the chamber \cite{soft1, soft2}. Their customizable, deformable nature and compliance make them suitable to biomedical applications as opposed to rigid and stiff mechanical robot components -- impractical in enabling articulation of human body parts. Our final design is a deformable IAB and a soft-robotic actuator specifically to address the problem deflection or attenuation of radiation beams.
	
	The paper is structured as follows: Section \ref{sec:sys_des} gives an overview of the system design and hardware set-up; Section \ref{sec:head_vision} details the vision algorithm used to determine the position of the patient's head; Section \ref{sec:sys_id} describes the identification of the soft robot model, and the control system design is presented in Section \ref{sec:cont_des}. Experimental results are presented in Section \ref{sec. expt} and we discuss future work and conclude the paper in Section \ref{conclusions}.
	
	\vspace{-0.2em}
	\section{System Design}\label{sec:sys_des}
\begin{figure}[tb]
		\centering
		\includegraphics[keepaspectratio = false, width=3.6in, clip=true, angle=0, center]{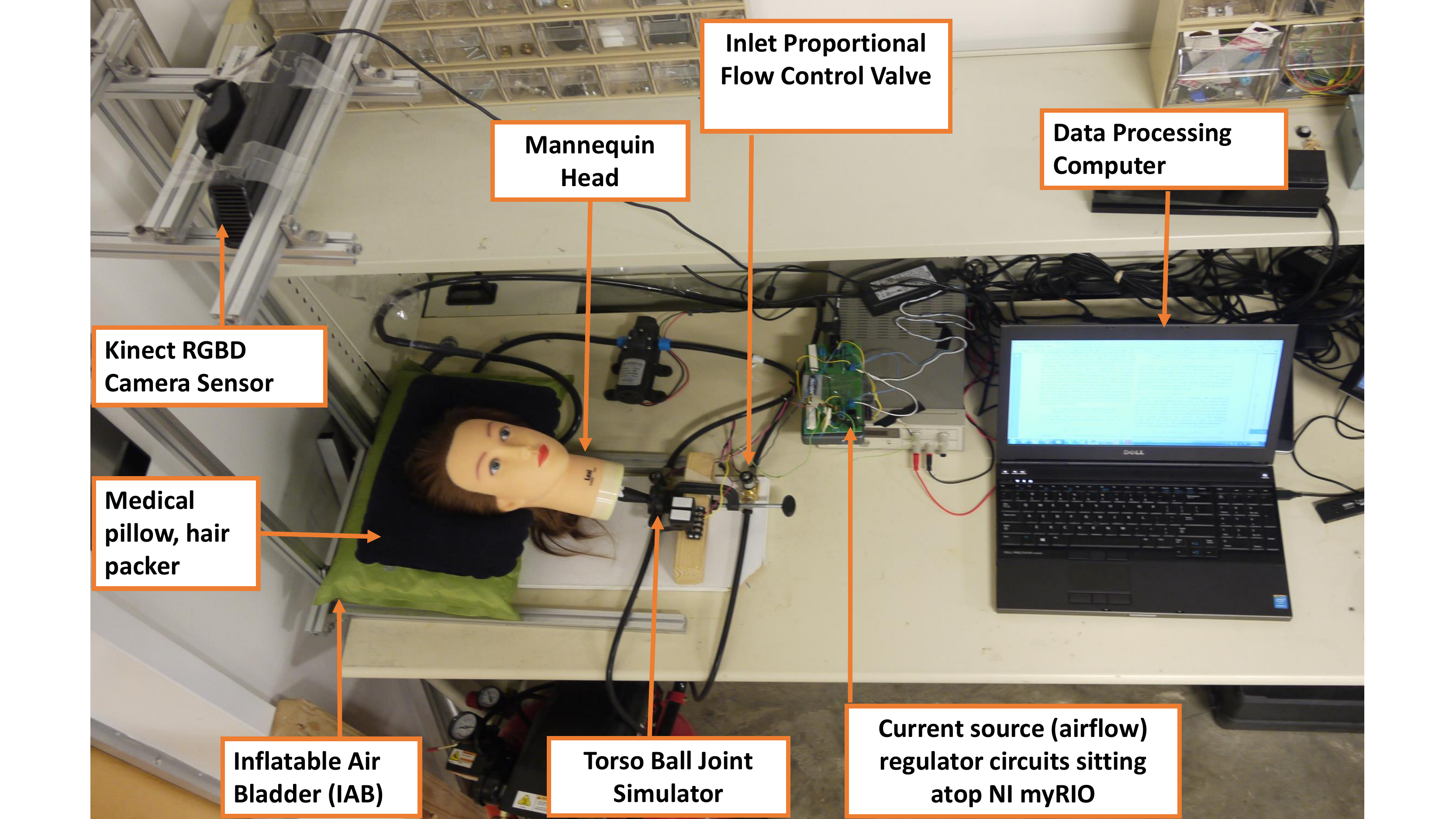}
		\vspace{0.1mm}
		\caption{Experimental Testbed}
		\label{Fig. 1.}
	\end{figure}

	\vspace{-0.5em}
	The system set-up is shown in Fig. \ref{Fig. 1.}. The patient simulator is a \/*human-like texture hair-cutting*/ Lexi cosmetology mannequin head (11" high, 6" wide) \/*made of protein fiber*/ with a hollow base that allows for a placeholder clamp. To simulate torso-induced neck motion, we attach a ball joint\/*neck placeholder in the hollow base of the head*/ in the hollow base of the head. The soft robot actuation mechanism combines a inflatable air bladder (19" x 12") made of lightweight, durable and deformable polyester and PVC, two current-controlled proportional solenoid valves (Model PVQ33-5G-23-01N, SMC Co., Tokyo, Japan), and a pair of silicone rubber tubes (attached to a T-port connector at the orifice of the IAB) in order to convey air in/out of the IAB. A 1HP air compressor supplied regulated air at 30 psi to the inlet actuating valve, while an interconnection of a 60W micro-diaphragm pump and a PVQ valve removed air from the outlet terminal of the IAB. The diaphragm pump creates the minimum operational differential pressure required by the outlet valve.
	
	We mount a Microsoft Kinect RGBD camera at approximately $710$mm above the manikin head, with the IAB fully deflated. A medical pillow was surrounds the head to reduce infra-red wavelengths scattering caused by the hair on the mannequin head, improve image processing for face extraction and negate undesirable head rotations. The vision algorithm was implemented on a 32GB RAM DELL Precision Laptop that ran 64-bit Windows 7.1 on an Intel Core i7-4800MQ processor. The real-time control processing was implemented on a National Instruments myRIO embedded system running LabVIEW 2014.
	%
	\vspace{-0.3em}
	\section{Vision-Based Head Position Estimation}   \label{sec:head_vision}
	The Kinect camera, though insufficient for clinical use, is reasonable for development and laboratory testing. H\&N Radiotherapy verification and validation experiments will incorporate the high-precision VisionRT 3D surface\footnote{\tiny \href{http://www.visionrt.com/products_solutions/alignrt}{Vision RT -- AlignRT Real-Time Patient Tracking, Patient Set-up in Radiation Therapy}} imaging system, approved for clinical use and capable of capturing a patient's position with the sub-millimeter spatial and sub-degree rotational accuracy. We use the near mode depth range of the Kinect sensor, i.e. 400mm -- 3000mm \cite{ms constants}, and the 640 $\times$ 480 depth image resolution and stream images at 30 frames per second.  We adopted the Microsoft Kinect SDK version 1.5.2 and OpenNI .NET framework \cite{c18, c19} for rapid prototyping of the experimental testbed.  
	
	An active appearance model (AAM) \cite{c14} was employed for face tracking, as it is a fast and robust method that uses statistical models of shape and gray-level appearance of faces. We adopted Smolyanski et al's approach \cite{c13}, which uses depth data to constrain a 2D + 3D AAM fitting. The approach in \cite{c13} was extended to a non-human object, i.e. the mannequin head in Fig. \ref{Fig. 1.}, by initializing the face tracker with a qualitatively determined region of interest. The face tracker utilizes both depth and color data but computes 3D tracking results in the video camera space. The video camera space is a right-handed system with the Z-axis pointing towards the face being tracked and the Y-axis pointing in the vertical direction.
	
	The points corresponding to the tip of the nose is fairly invariant to movement of facial muscles. Therefore, the Z-coordinates of points corresponding to the nose area were averaged, and this was used to determine the patient position with respect to the origin of the camera frame. We mapped this result to world space, i.e. the heads displacement above the table using the relation
	\vspace{-0.2em}
	\begin{eqnarray}
	y(t) = y_m - y_h    \label{eq:depth}
	\end{eqnarray}
	where $y(t)$ is the displacement of the head from the table; $y_m$ is the head displacement as measured by the camera \/*during the IAB inflation or deflation*/; $y_h$ is the mounting height of the camera above the table.
	\begin{figure}[tb]
		\centering
		\centering
		\includegraphics[trim=0mm 0mm 10mm 0mm, height=1.5in, clip=true]{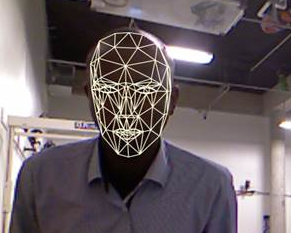} 
		\hspace{0.05cm}
		\includegraphics[trim=5mm 0mm 5mm 15mm, height=1.5in, clip=true]{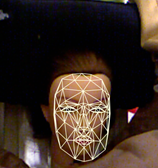} 
		\vspace{0.2cm}
		\caption{Depth Constrained 3D Face Tracking of a human head (left) and a mannequin head (right) using AAM.} 
		\label{Fig. 2}
	\end{figure}
	
\begin{figure}[tb]
		\centering	
		\begin{tikzpicture}[scale=4, node distance = 1.4cm, auto]
		\node [block] (Midpoint) {Retrieve midpoint of depth coordinates};			
		\node [block, left of =	Midpoint, node distance=4cm] (init) {Initialize Kinect and Generate Depth Map};
		\node [block, below of=Midpoint, node distance=1.6cm] (send) {Send coordinates to myRIO via UDP packets};
		\node [decision, below of=send, node distance=2.2cm] (decide) {Missing Frame Data?};
		\node [block, left of=decide, node distance=4.5cm] (Yes) {Use previous frame's depth data};
		\node [block, below of=decide, node distance=2.1cm] (no) {Apply FIR point-by-point filter};
		\node [block, below of=no, node distance=1.6cm] (Use) {Pass on measurement data to control patient head};
		%
		\path [line] (init) -- (Midpoint);
		\path [line] (Midpoint) -- (send); 
		\path [line] (send) -- (decide); 
		\path [line] (decide) -- (no); 
		\path [line] (decide) -- node [near start, color=black] {Yes} (Yes);
		\path [line] (decide) -- node [near start, color=black] {No} (no);
		\path [line, dashed] (Yes) |- (no);
		\path [line] (no) -- (Use);	
		\end{tikzpicture} 
		\vspace{0.2cm}
		\captionof{figure}{Vision Flowchart Using the OpenNI .NET Assembly}
		\label{Fig. 3.}
	\end{figure}
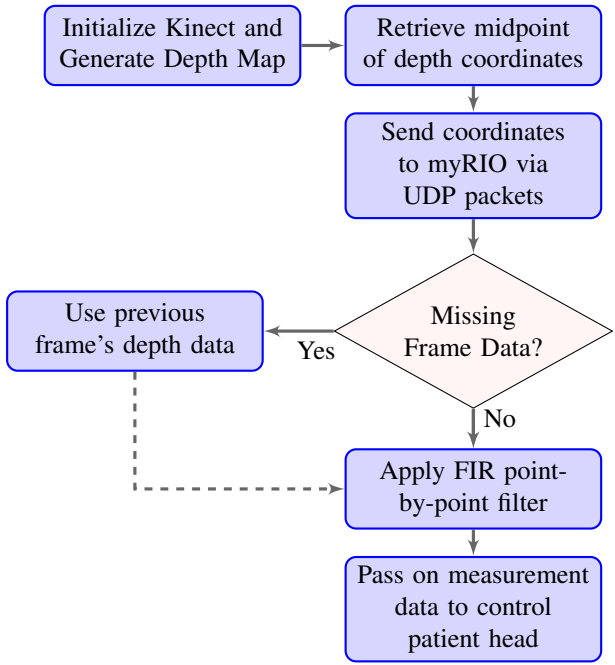

	The tracked head position value from \eqref{eq:depth} was transferred from the vision processing workstation to myRIO over a local wireless network using the user datagram protocol (UDP). We chose UDP over other handshaking, dialog-based connection transmission models because the application is a real-time sensitive one. The typical problem of dropped packets with UDP-based connections, is preferable for our goals over delayed packets, which can occur in other connection-based protocols. An algorithm for Network interface level error-checking and correction was handled in myRIO using the procedure described in Fig. \ref{Fig. 3.}.
	
	A deterministic protocol was implemented on myRIO to prioritize transmission of Kinect measurement data and to eliminate synchronization errors between the server and the client -- a common issue with the Windows operating system. To ensure the deterministic task does not monopolize other myRIO processor resources, a timing engine was employed with an execution rate equal to that of the depth image processing loop on the Windows workstation, i.e. 30Hz. Finally, a 20th order nonrecursive point-by-point finite-impulse response filter was employed to mitigate measurement noise from streamed data.
	\/*
	\eqref{eq.filter}.
	\begin{equation}
	y_i = \sum_{k=0}^{n-1} h_k x_{i-k}		\label{eq.filter}
	\end{equation}
	where $x$ is the input sequence to the filter; $y$ is the filtered sequence; $n$ is the number of filter coefficients, in this case 20; and $h$ are the filter coefficients.
	*/
	
	\section{System Identification and Modeling}\label{sec:sys_id}
	A reliable system model is necessary to design a stable controller with required time and precision characteristics. 
	
	\subsection{Data Collection}
	With the regulated air canister providing a constant pressure of 30 psi, a periodic, persistently exciting input current in the form of a sawtooth waveform was used to excite the inlet PVQ valve such that the experiment was open-loop informative \cite[p. 414]{c22}. Airflow out of the outlet valve was kept constant by opening it to the mid-position of its operating range. This varied the head position through an open-loop inflation/deflation process of the IAB. 
	
	The current to the inlet valve, $u(t)$, was band-limited such that it had no power above 10Hz, .i.e., the Nyquist frequency of the valves, and its spectrum coincided with the spectrum of the  discrete time signal.  The output signal, $y(t)$, is the height of the head given by \eqref{eq:depth}.
	We acquired 8,800 samples of the input and output signals for data modeling, and a second set of 8,800 samples was collected for model validation. 
	\subsection{Data Pre-Processing and System Model Identification}
	Consider a single input, single output relationship in the form of a linear difference equation
	\vspace{-0.12cm}
	\begin{align}
	\begin{split}
	y(t)+a_1y(t-1)+\cdots+a_ny(t-n) &=  \\
	\ \qquad b_1u(t-1)+\cdots+b_mu(t-m)     \label{eq. 2.0}
	\end{split}
	\end{align}
	Rewriting (\ref{eq. 2.0}) 
	such that it models a one-step-ahead predictor, we have
	\begin{eqnarray}
	\begin{split}
	y(t) &=-a_1y(t-1)-\cdots-a_ny(t-n)+ \cdots\\
	&\qquad \qquad \qquad \quad \hspace{0.68cm} + b_1u(t-1)+ b_mu(t-m).		\label{eq. 2.01}
	\end{split}
	\end{eqnarray}
	We want a model structure from the collected data set, $Z^N = \{u(1), y(1), \cdots, u(N), y(N)\}$, parametrized by mapping from the set of all past inputs and outputs, $Z^{t-1}$, to the space of the model outputs.  Denote the model as $\hat{y}(t|\theta)$
	\begin{align}
	\hat{y}(t|\theta) & = g(\theta, Z^{t-1})          \label{eq. 2.1}
	\end{align}
	where $\theta$ is the set of estimated coefficients to satisfy \eqref{eq. 2.0}
	\begin{eqnarray}
	\theta &= \begin{bmatrix}
	a_1 & \cdots & a_n & b_1 &\cdots & b_m &
	\end{bmatrix}^T.
	\end{eqnarray} 
	The identification goal is to identify the best model in the set, $Z^N$, guided by frequency distribution analysis. Removing means and linear trends in collected data will minimize the effects of disturbances that are above the frequencies of interest to system dynamics, and will eliminate occasional outliers and non-continuous records in collected data \cite[Ch. 3, pp. 414]{c22}. Therefore, acquired data was normalized using 
	\begin{align}
	u_{ave}(t) = u(t) - \bar{u}, \qquad y_{ave}(t) = y(t) - \bar{y} \label{eq3}
	\end{align}
	where $\bar{u} = \dfrac{1}{N}\sum\limits_{t=1}^N u(t)$ and $\bar{y} = \dfrac{1}{N}\sum\limits_{t=1}^N y(t)$ are the corresponding sample means, $n$ is the discrete time index and $N$ is the total data length \cite[Ch. 1\/*, pp. 13*/]{c22}. 
	Linear trends were then removed using
	\begin{align}
	u_{d}(t) = u_{ave}(t) -A\theta_u, \qquad y_{d}(t) = y_{ave}(t) -A\theta_y \label{eq4}
	\end{align}
	where $\theta_u$ and $\theta_y$ are the solutions to the least-square fit equations
	\begin{align}
	(A^TA)\theta_u = A^Tu, \qquad (A^TA)\theta_y = A^Ty  \label{eq5}
	\end{align}  
	and
	\begin{align}
	A^T = \begin{bmatrix}
	1&  1 & \cdots & 1 & 1\\ 
	\frac{1}{N}& \frac{2}{N} & \cdots & \frac{N-1}{N} & 1   \label{eq6}
	\end{bmatrix} .
	\end{align}
	\begin{figure}[tb]
		\centering
		\includegraphics[keepaspectratio = true, width=3.6in, clip=true]{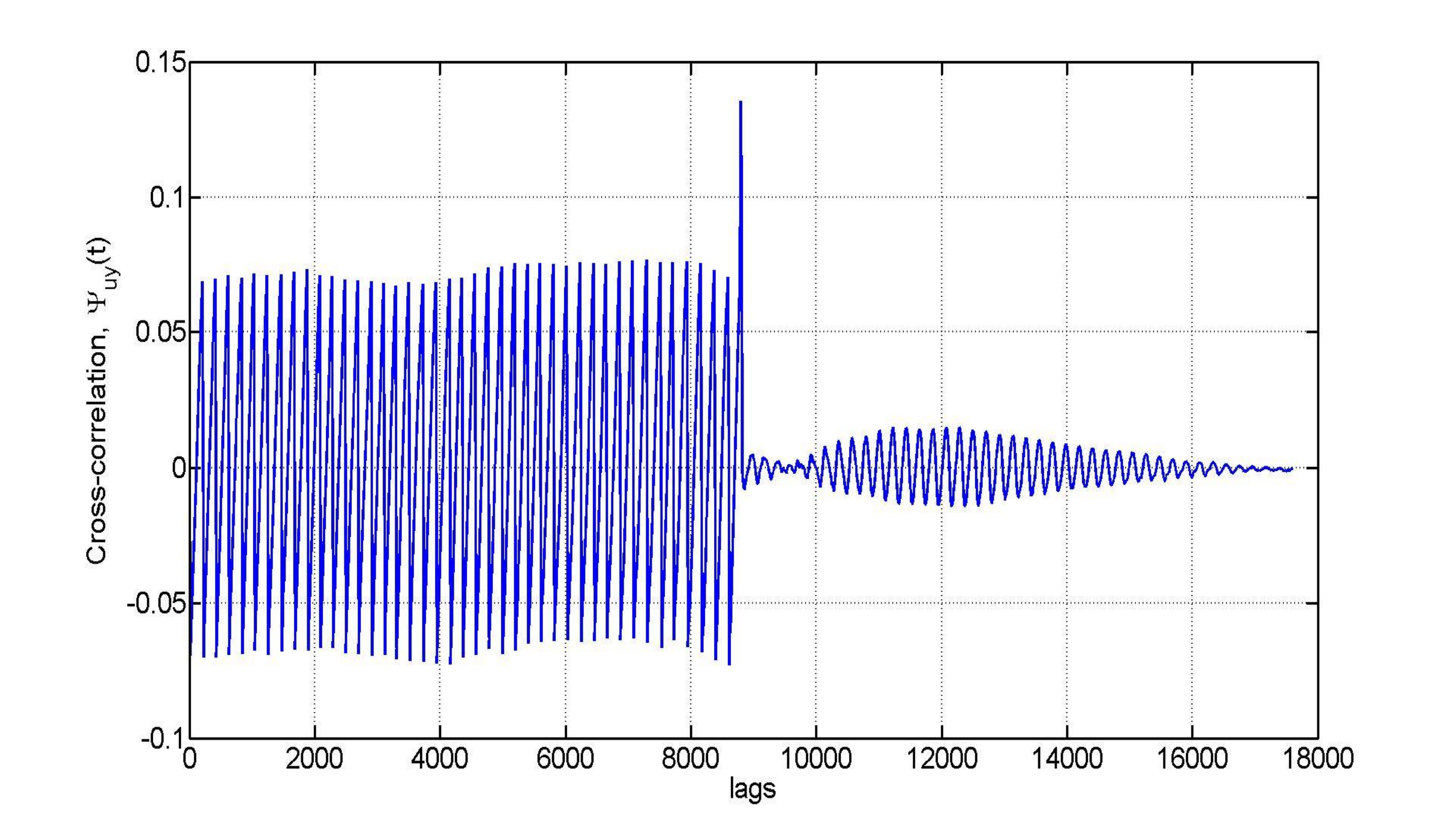}
		\caption{CCF of Input and output Signals}
		\label{Fig. 5.5}
	\end{figure}
	
	To examine, the relationship between the input and output signals, the normalized cross-correlation function (CCF) between $u(t)$ and $y(t)$, was determined as
	\begin{align}
	\psi_{uy}(\tau) &= \dfrac{\sum\limits_{t=\tau+1}^N\left[u(t-\tau) - \bar{u}\right]\left[y(t)-\bar{y}\right]} {\sqrt{\sum\limits_{t=1}^N\left[u(t) - \bar{u}\right]^2}\sqrt{\sum\limits_{t=1}^N\left[y(t) - \bar{y}\right]^2}} \label{eq7}
	\end{align} 
	\begin{center}
		$\tau = 0, \pm1, \cdots, \pm (N-1)$.
	\end{center} 
	%

	\begin{figure}[tbph]
		\centering
		\includegraphics[trim=0mm 10mm 0mm 40mm, width= 3.2in, clip=true]{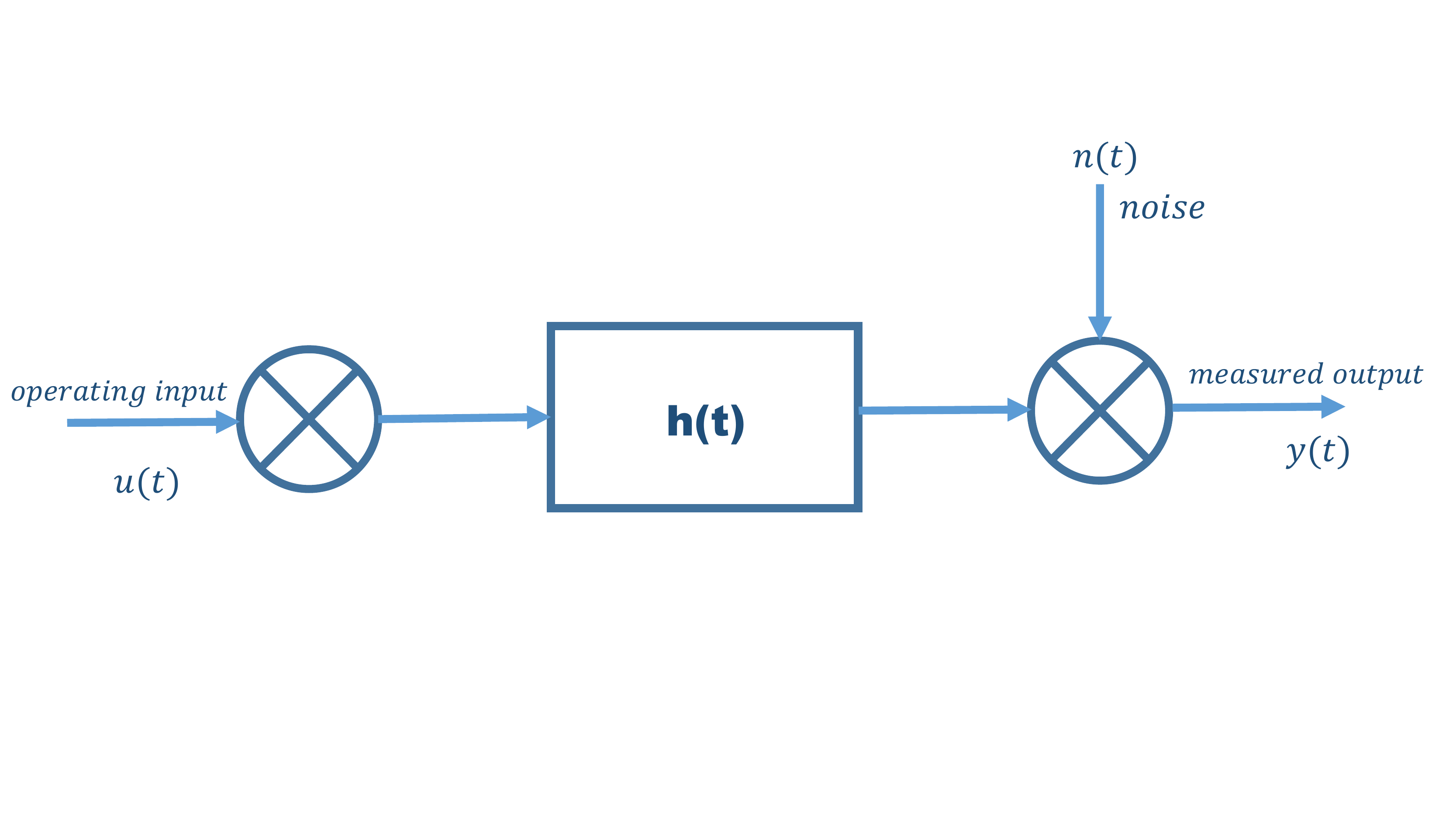}
		\vspace{-1.1cm}
		\caption{Impulse Response Correspondence with the CCF}
		\label{Fig. wiener.}
	\end{figure}
	
	\vspace{-0.3cm}
	Since the CCF is the convolution of the system impulse response, $h(t)$ and the process auto-correlation function, $\phi_{uu}(t)$, the Wiener-Hopf equation \eqref{eq7} can be rewritten as 
	\begin{dmath}
		\psi_{uy}(\tau)=\int h(\nu) \mathbb{E}[u(t)u(t+\tau - \nu)]d\nu = \int h(\nu) \psi_{uu}(\tau - \nu)d\nu	\label{eq. wiener-hopf}
	\end{dmath}
	where $\mathbb{E}$ denotes the expectation operator. Equation \eqref{eq. wiener-hopf} implies the CCF between the output and test input is proportional to the system impulse response when the input is a white noise signal\cite[p. 13]{c22}.
	The CCF in Fig. \ref{Fig. 5.5} is not a correct estimate of the system impulse response, since the excitation input was not a white noise sequence. Therefore, the input and output were prewhitened with a white noise input sequence, $ u_{pw}(t) = u(t)F(z^{-1})$, where $u_{pw}(t)$ is a zero-mean white input  sequence and $F(z^{-1})$ is an autoregressive filter of order 20 defined as
	\begin{align}
	F(z^{-1}) & = 1 + \varsigma_1 z^{-1}+\varsigma_2 z^{-2}+\cdots+\varsigma_{20}z^{-20}. \label{eq9}
	\end{align}
	The parameters $\varsigma_i$, ($i = 1, 2, \cdots, 20$) were estimated by fitting an autoregressive model to $u(t)$ and were generated with the `\texttt{ar}' command in MATLAB. 
	%
	
	The estimation result after fitting the white noise auto-regressive model was found to have 76.75\% fit to estimation data, with a mean squared error (MSE) of 78.11 mm$^2$. 
	The normalized auto-correlation function tells of the filter quality ${F}(z^{-1})$ and is given by
	\begin{dmath}
		\psi_{uu}(\tau) = \dfrac{\sum\limits_{t=1}^N\left[u(t) - \bar{u}\right]\left[u(t+\tau)-\bar{u}\right]} {\sum\limits_{t=1}^N\left[u(t) - \bar{u}\right]^2} \,	 	\label{eq.acf}
	\end{dmath}
	where $\tau = 0, \pm 1, \cdots, \pm (N-1)$.
	
\begin{figure}[tb]
		\centering
		\includegraphics[ width=3.6in, keepaspectratio = true, clip=true]{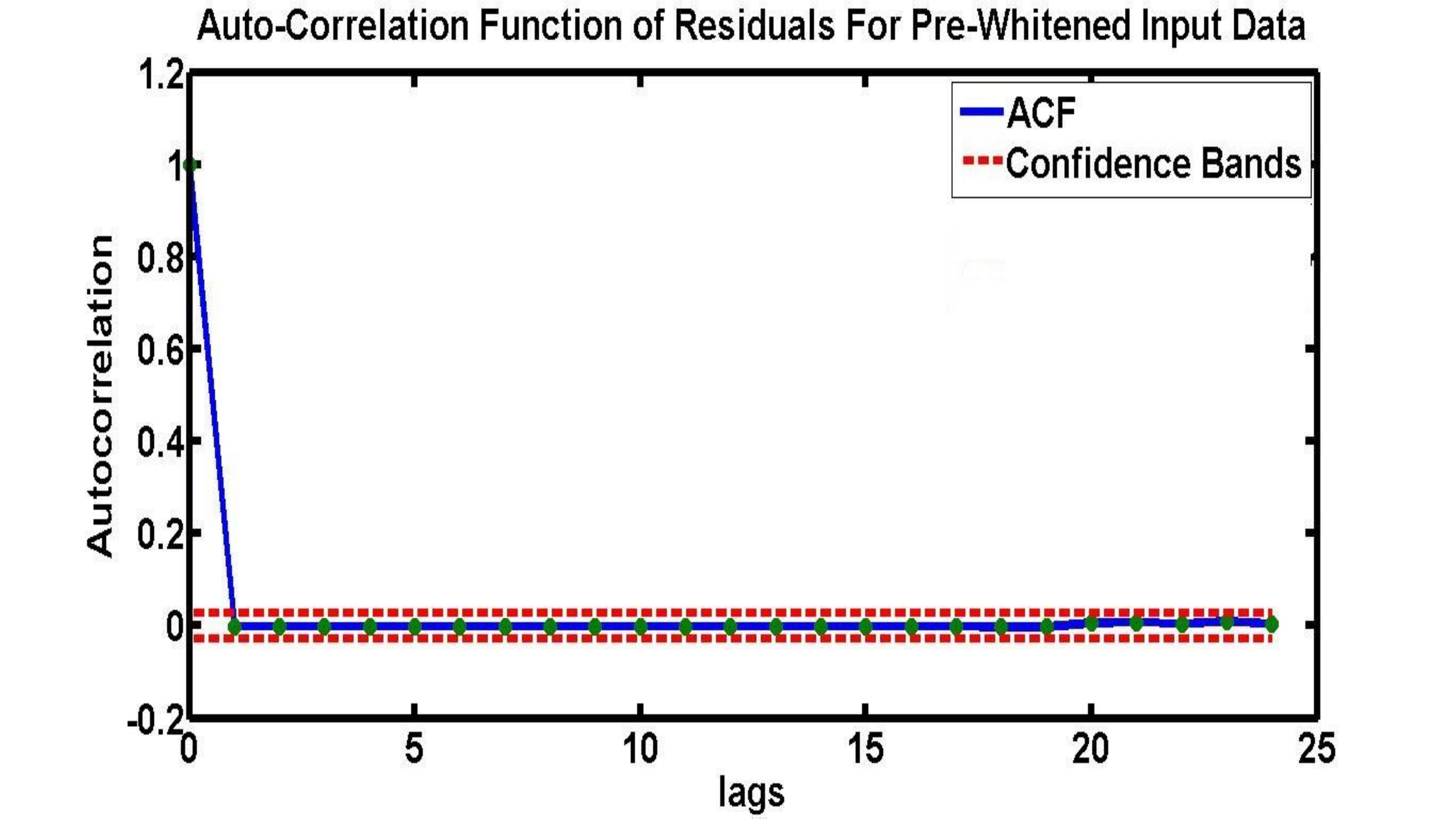}
		\caption{Correlation Function of Residuals}
		\label{Fig. 10.}
	\end{figure}

\begin{figure}[tbph]
		\centering
		\includegraphics[keepaspectratio = true, width=3.6in, clip=true]{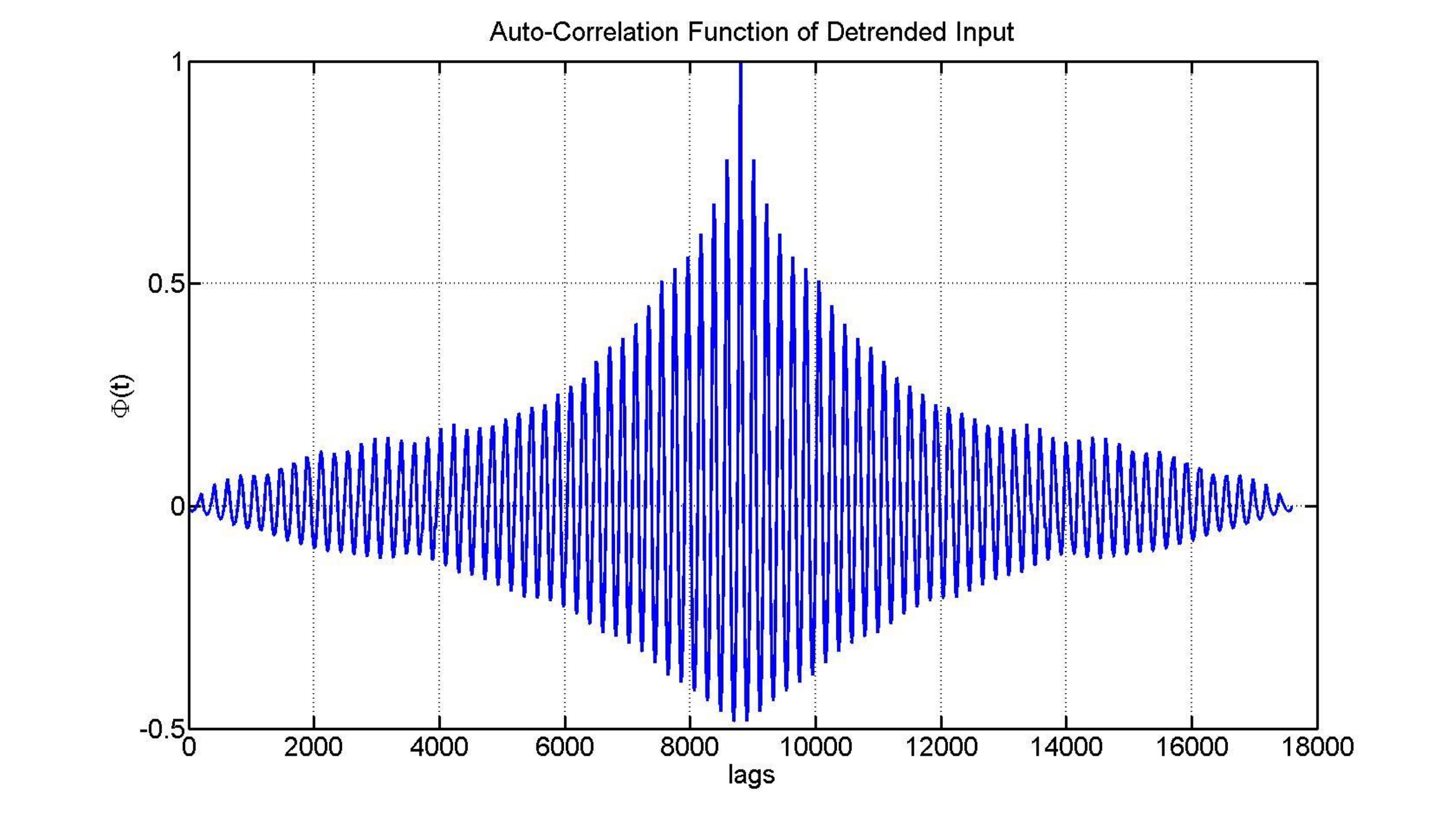}
		\caption{ACF of Input and output Signals}
		\label{Fig. acf.}
	\end{figure}

	The auto-correlation function of the residuals of the pre-whitened input signals, seen in Fig. \ref{Fig. 10.}, are within 95\% confidence bands (the dashed red lines). Hence, we conclude that we correctly estimated the filter.\/*
	Also, the correlation function of Fig. \eqref{Fig. acf.} shows the impulse response after pre-whitening resembles that of a second-order function.
	*/
	\/*
	*/
	To find an optimal sub-model for the identified system that will tolerate nonlinearities and handle disturbances well, our final choice was a linear, second-order grey-box process model on the detrended data with quality measurable by the MSE. This model choice is informed by the previous impulse response analysis which suggests a delay in the system (Fig. \ref{Fig. 10.}) and gave an affordable model cost acceptable for solving $\hat{\theta}_N$. A high-order complex model may be marginally better but may not be worth the higher cost \cite[\S16.8]{c22}.
	
\begin{figure}[tb]
		\centering
		\includegraphics[trim=15mm 0mm 15mm 0mm, width=3.2in, clip=true]{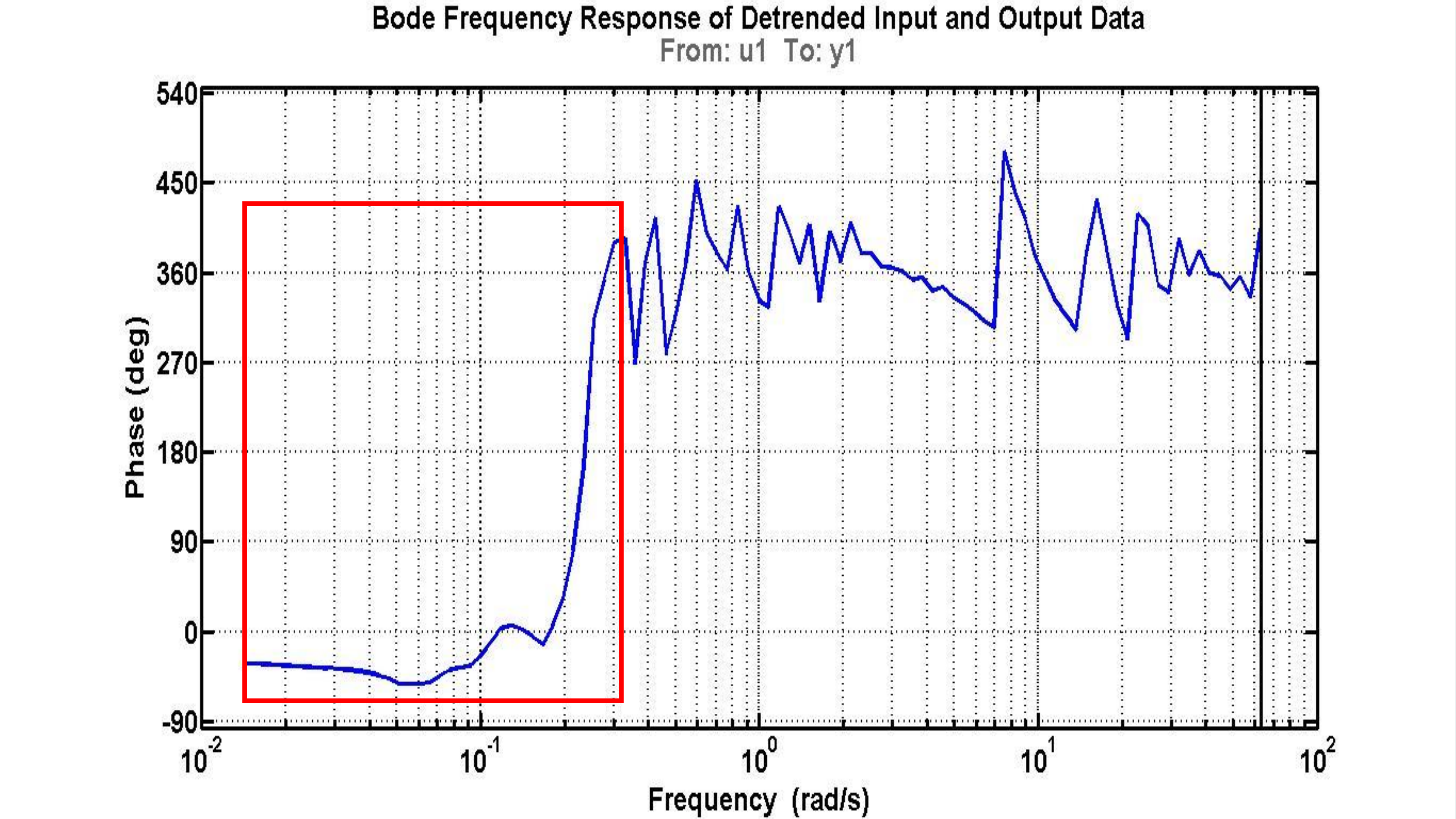} 
		\caption{Power Spectral Density Phase Plot of Detrended Input and Output Signals}
		\label{Fig. 13.}
	\end{figure}

	By analyzing the bode response of the spectral frequency density distribution of the detrended data, we chose the approximately linear frequency range (0.00232 rad/sec -- 6.85 rad/sec) in the frequency distribution of Fig. \ref{Fig. 13.} to represent the desired model.
	\subsection{Model Estimation}
	Using term selection and parameter estimation, we fit a second-order process model with transfer function form
	\begin{align}
	G(s) & = K_p \dfrac{1+sT_z}{(1+sT_{p_1})(1+sT_{p_2})}e^{-sT_d}.   \label{eq15}
	\end{align}
	where $T_{z}$, $T_{p_1}$, and $T_{p_2}$ are respectively the process zero and the time constants contributed by the first and second pole of the system; $K_p$ is the process dc gain, and $T_d$ is the process dead time. The delay is a result of the non-collocation of the sensor and actuator. The identified parameters of (\ref{eq15}) are listed in Table \ref{table:ident}. 
	
	\begin{savenotes}
		\begin{table}[tbph]
			\centering
			\caption{Parameter Estimation Results for Soft Robot System}
			\begin{tabular}{|c|c|c|c|c|c|c|}
				\hline \rule[-2ex]{0pt}{5.5ex} $K_p$   & $T_z$    & $T_{p_1}$ & $T_{p_2}$  & $FPE$  &  $MSE$  & $T_d$\\ 
				\hline \rule[-2ex]{0pt}{5.5ex}  1.0015 & -0.58354 & 100       &  9.7257    & 1.672 & 0.05498 & 2 \\ 
				\hline  
			\end{tabular}
			\label{table:ident}
		\end{table}
	\end{savenotes}
	
	A first-order measurement noise component ARMA disturbance model, has been fit into $G(s)$ 
	\begin{align}
	y(s) = G(s)u(s) + \dfrac{C(s)}{D(s)}e(s),			\label{eq16}
	\end{align}  where $e(s)$ is a white noise, $C(s)= s + 899.3$, and $D(s) = s + 7.789$ . A prediction focus was used to weigh the relative importance of how closely to fit the data in the various frequency ranges. This favored the fit over a short time interval \cite[Ch. 3, Sec. 3-38]{sysidmatlab}.
	The model has 87.35\% fit to original data with improved quality as the final prediction error (FPE) and MSE shows.
	\/*
	\begin{align}
	G(s) = \dfrac{-0.0006\left(s-1.7137\right)}{\left(s+0.01\right)\left(s+0.1028\right)}exp^{-2s}.    \label{PlantTF}
	\end{align}
	*/
	\subsection{Residual Analysis} 
	To verify the model accuracy with respect to our control goal, we employed canonical analysis by computing the prediction errors as a frequency response from the inputs to the residuals not picked up by the model.
	Defining the outputs predicted by the model as $\hat{y}(t|\hat{(\theta)}_N$, the errors from the modeling process are the residuals
	\begin{align}
	\alpha(t)=\alpha(t, \hat{\theta}_N)=y(t)-\hat{y}(t|\hat{(\theta)}_N  .   \label{eq. 18}
	\end{align}
	A basic statistics for the residuals from the model such as
	\begin{equation}
		S_1 = \text{max}_t |\alpha(t)|, \qquad S^2_2 = \frac{1}{N}\sum_{t = 1}^{N}\alpha^2(t)
	\end{equation} 
	will inform us about the model's quality since the upper limit of $S_1$ or the average error of $S_2$ for all data we have will also be bound for all future data. In order to check that the model would work for a range of possible inputs, we study the covariance between residuals and past inputs
	\begin{equation}
		\hat{R}^N_{\alpha u} (\tau)=\frac{1}{N}\sum_{t=1}^{N}\alpha(t)u(t-\tau)
	\end{equation}
	and deem the model is invariant to other inputs if the numbers, $\hat{R}^N_{\alpha u} (\tau)$, are small enough so that $y(t)$ could not have been better predicted, i.e., there is no part of $y(t)$ not picked up y the model $G(s$)
	We compare the estimates of the obtained linear model with the corresponding standard deviation (from the validation data set, $Z_v^N$) in Bode plots with estimated variance translated to confidence intervals.
	\begin{figure}[tb]
		\centering
		\includegraphics[trim=0mm 0mm 0mm 0mm, width=3.5in, clip=true]{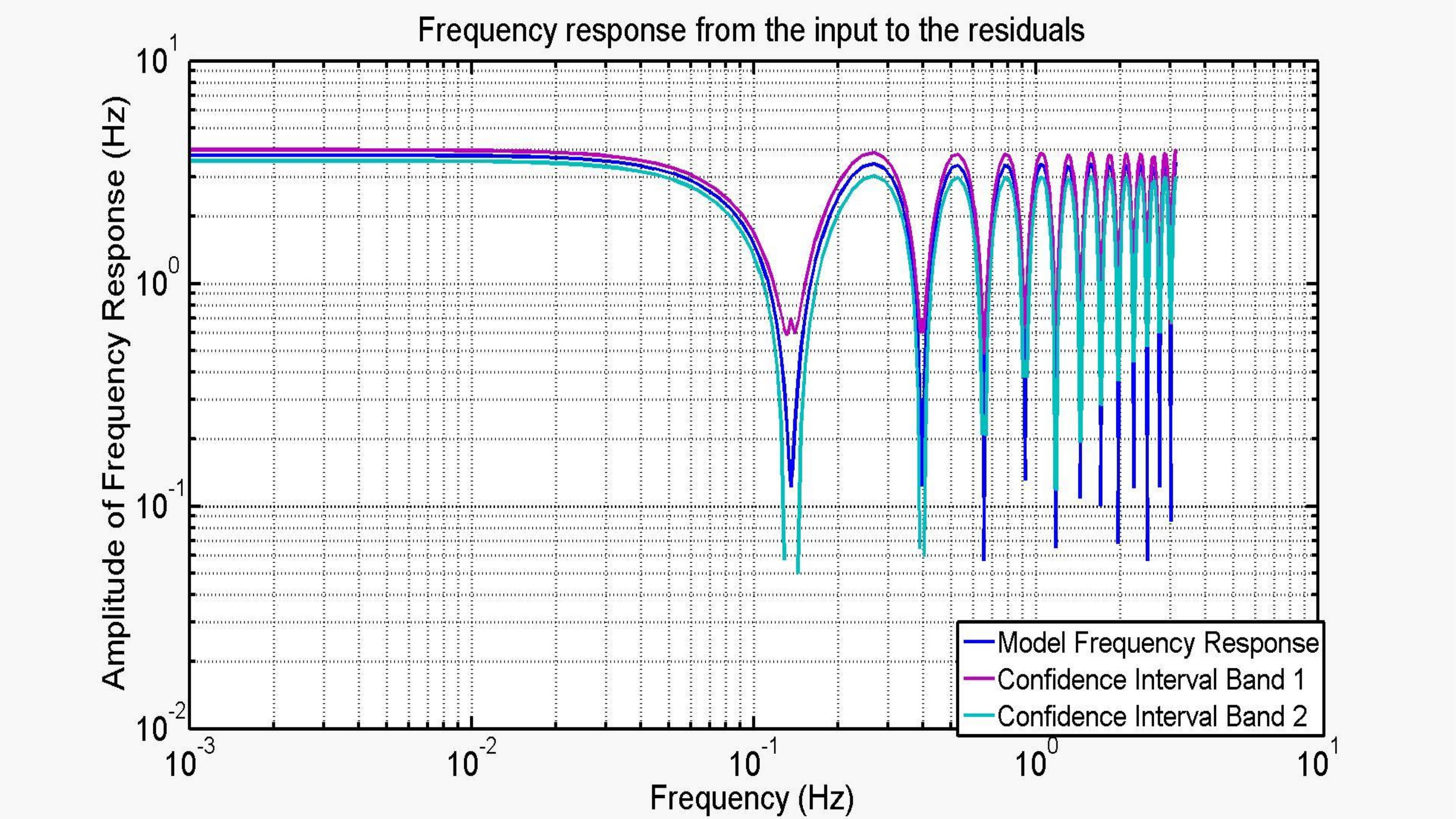}
		\caption{Frequency analysis from past inputs to residuals}
		\label{Fig. model error model.}
	\end{figure} 
	We see from Fig. \eqref{Fig. model error model.} that the model's frequency response generally stays within the 99\% confidence bands (the pink and purple lines), and conclude we have a reliable model. 
	\section{Control Design}\label{sec:cont_des}
\begin{figure}[tb]
		\centering 
		\includegraphics[trim=30mm 0mm 30mm 0mm, width=3.2in, clip=true]{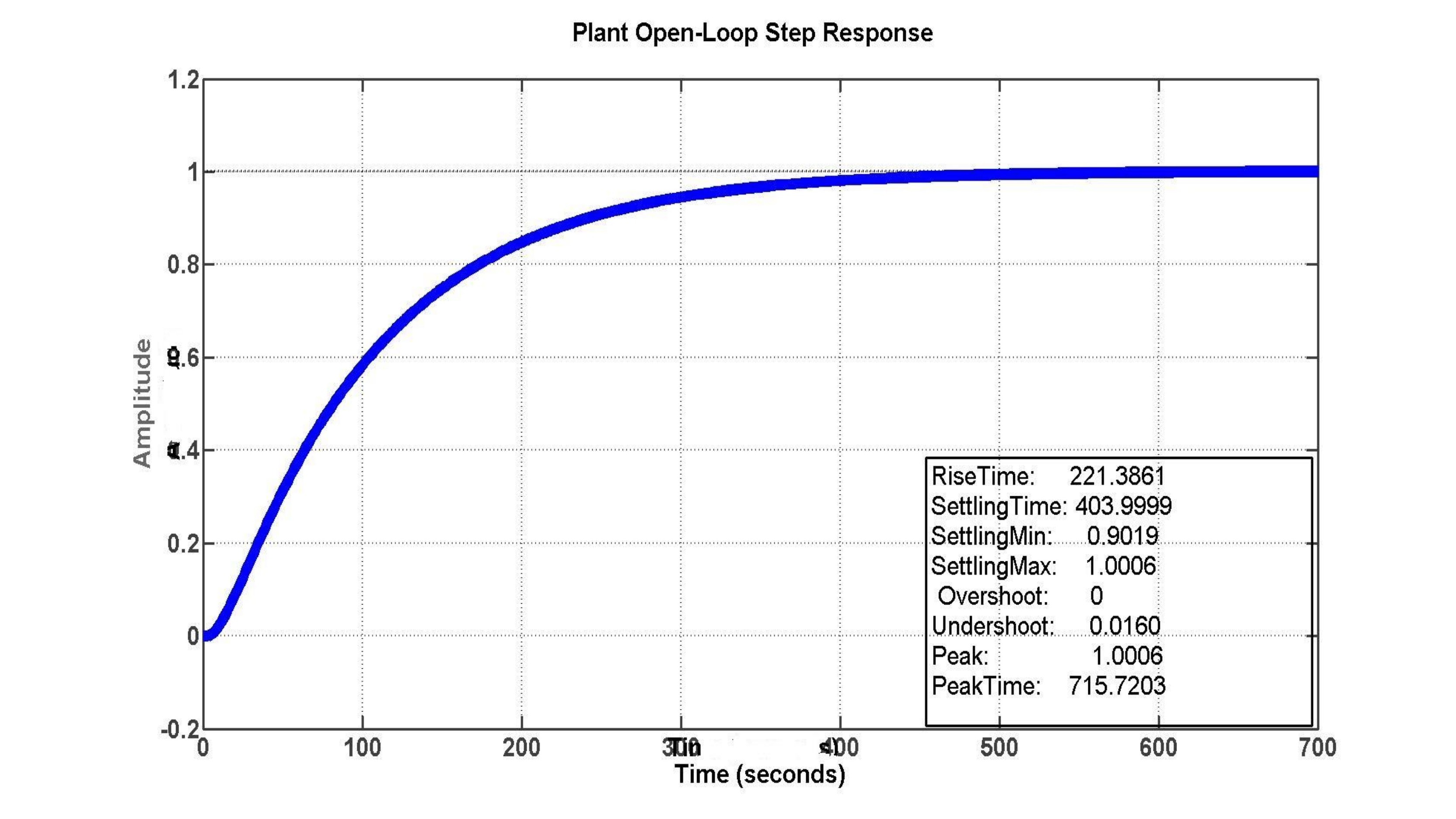} 
		\caption{Open-Loop Step Response of Identified Model}
		\label{Fig. 17.}
	\end{figure}

\begin{figure}[tb]
		\centering  
		\includegraphics[trim=30mm 0mm 30mm 0mm, width=3.2in, clip=true]{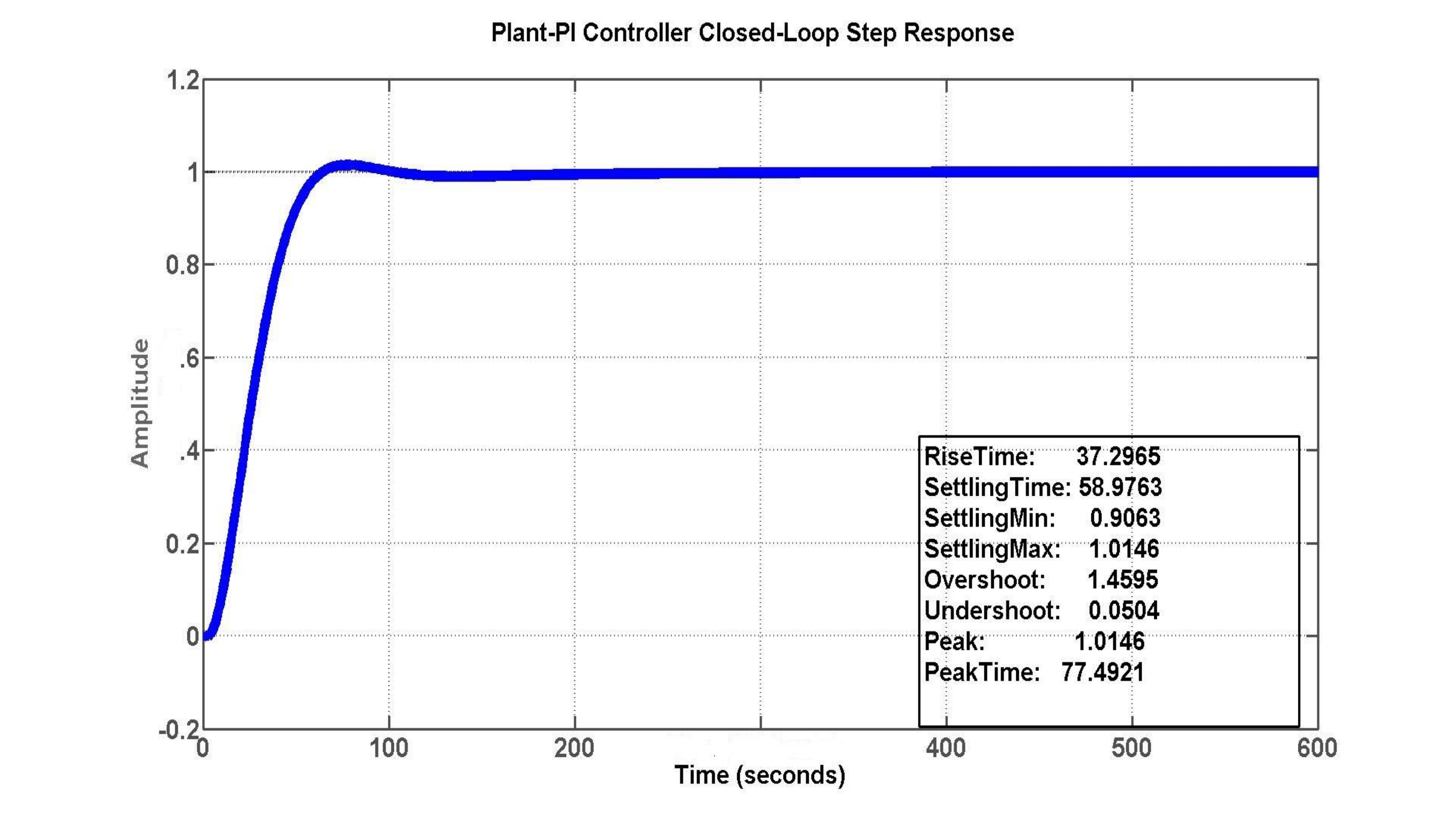}
		\caption{Closed-Loop Step Response of PI tuned Soft Robot System}
		\label{Fig. 18.}
	\end{figure}

	The step response of the open loop system (Fig. \ref{Fig. 17.}) shows the system is stable, but with a very slow transient response. We require a controller that gives closed loop stability and achieves a clinically acceptable response time (15 - 30 seconds) while balancing the trade-off between aggressiveness and robustness. To do this, a pole  was added at the origin and a zero was kept close to the introduced pole as in Fig. \ref{Fig. Model.} using the following PI-controller in a feedforward configuration with obtained model.
	\begin{align}
	G_c &= 3.79 + \dfrac{0.0344}{s}.  \label{eq.controller}
	\end{align}
	This reduced steady state error while maintaining transient characteristics.
	The closed-loop unit step response with the added controller is shown in Figure \ref{Fig. 18.}. The system's transfer function with the added PI controller of (\ref{eq.controller}) is
	\begin{align}
	\begin{split}
	G_{ol} &= \dfrac{-0.00228\left(s+0.009073\right)\left(s-1.7137\right)exp^{-2s}}{s\left(s+0.01\right)\left(s+0.1028\right)}  \label{eq. CLTF}
	\end{split}
	\end{align}
	where the delay was approximated with a second-order Pad\'{e} approximant of the form
	\begin{align}
	H(s) & = \dfrac{s^2 - 3s +3}{s^2 + 3s +3}.   \label{eq.pade}
	\end{align}
	This preserved the transient characteristics by sufficiently approximating the delay according to our control goal.
	The overall desired transient and frequency response was then realized with a feedforward PID-controller in series with the closed loop network of the PI-controlled soft robot system (\ref{Fig. Model.}). The PID controller, given by
	\begin{align}
	G_{PID}=3.4993 + \dfrac{0.054765}{s} + 55.8988s,          \label{eq.pd control}
	\end{align}
	corrected fluctuations in air flow into the IAB and improved the system's dynamic performance such that the overall closed loop network has the step response seen in Fig. \ref{Fig. 20.}.
	\/*
	\begin{figure}[tb]
		\centering 
		\includegraphics[trim=30mm 0mm 30mm 0mm, width=3.2in, clip=true]{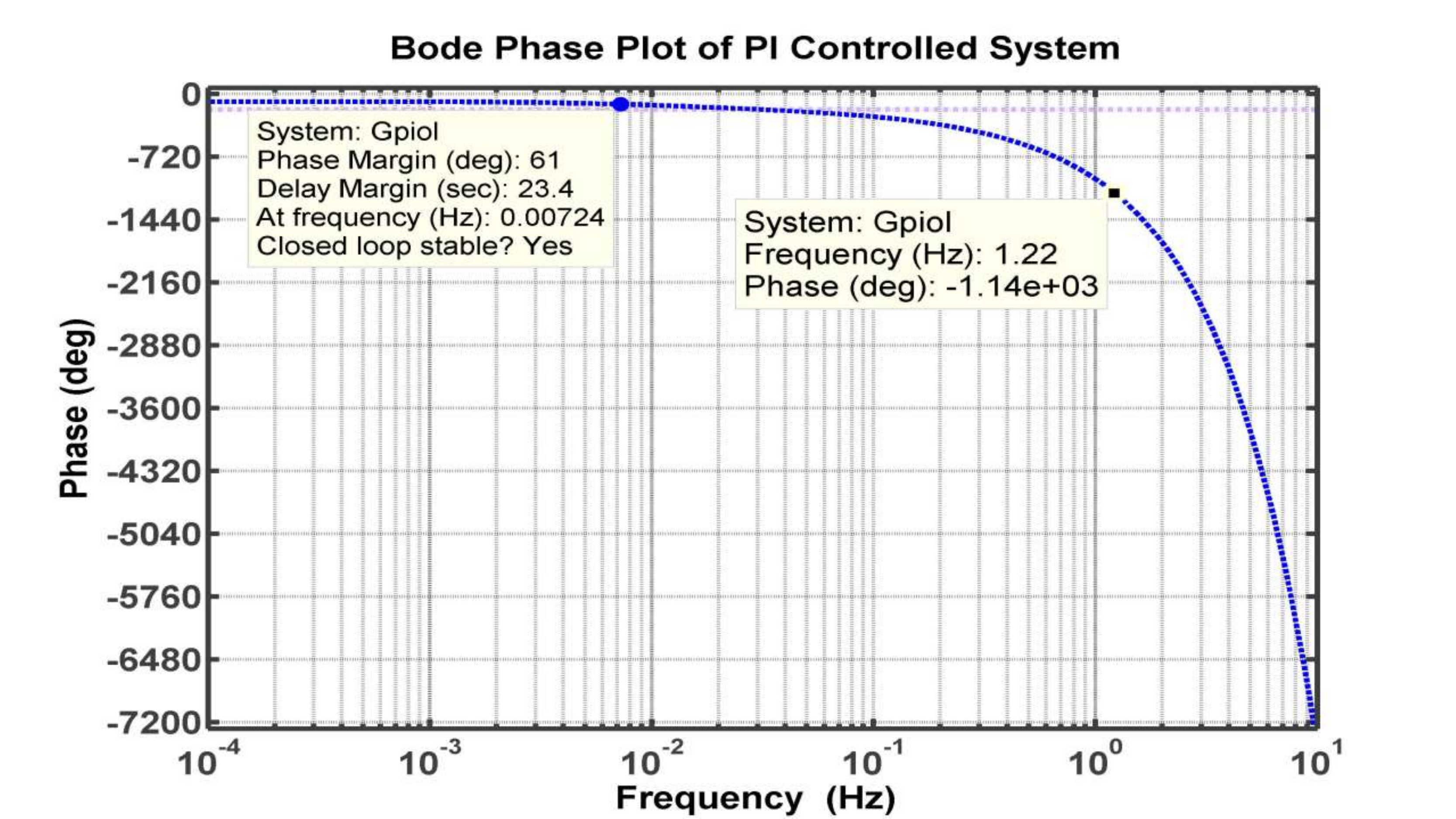} 
		\caption{Bode Phase Plot of PI-tuned system}
		\label{Fig. 19.}
	\end{figure} 
	*/
	\begin{figure}[tb]
		\centering
		\includegraphics[trim=30mm 0mm 30mm 0mm, width=3.2in, clip=true]{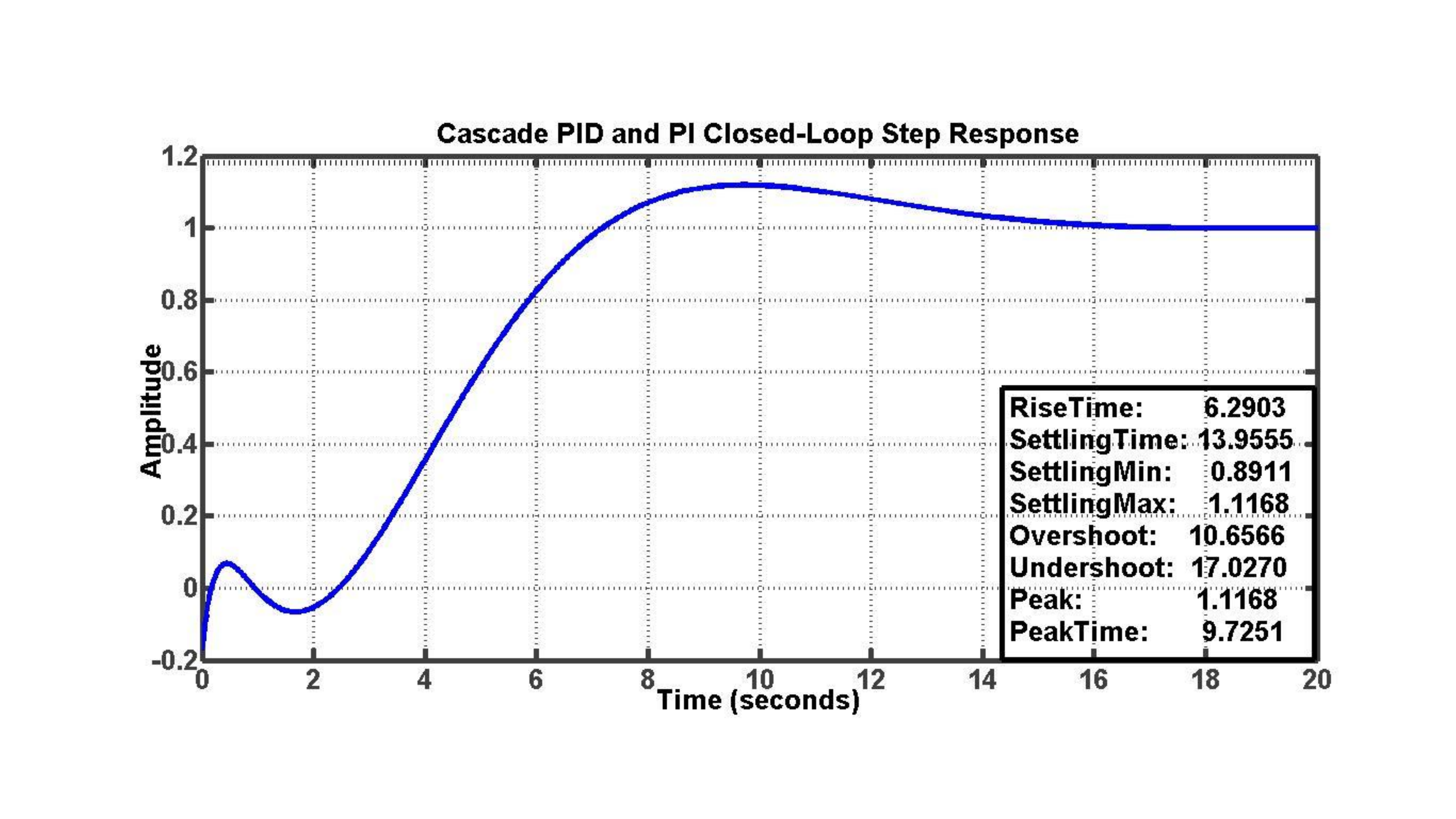}
		\caption{Closed Loop Step Response Plot of PID and PI Cascade Network}
		\label{Fig. 20.}
	\end{figure} 
	This produced a non-minimum phase system with settling time of approximately 14 seconds. As seen in Fig. \ref{Fig. 20.}, the system converges to steady state with a rise time of 6.29 seconds. The overall PID-PI control network (shown in Fig \ref{Fig. Model.}) is closed loop stable as the Bode plot of Fig. \ref{Fig. 21.} shows.
	\begin{figure}[tb]
		\centering
		\includegraphics[trim=0mm 20mm 0mm 55mm, keepaspectratio = true, width=3.4in, clip=true]{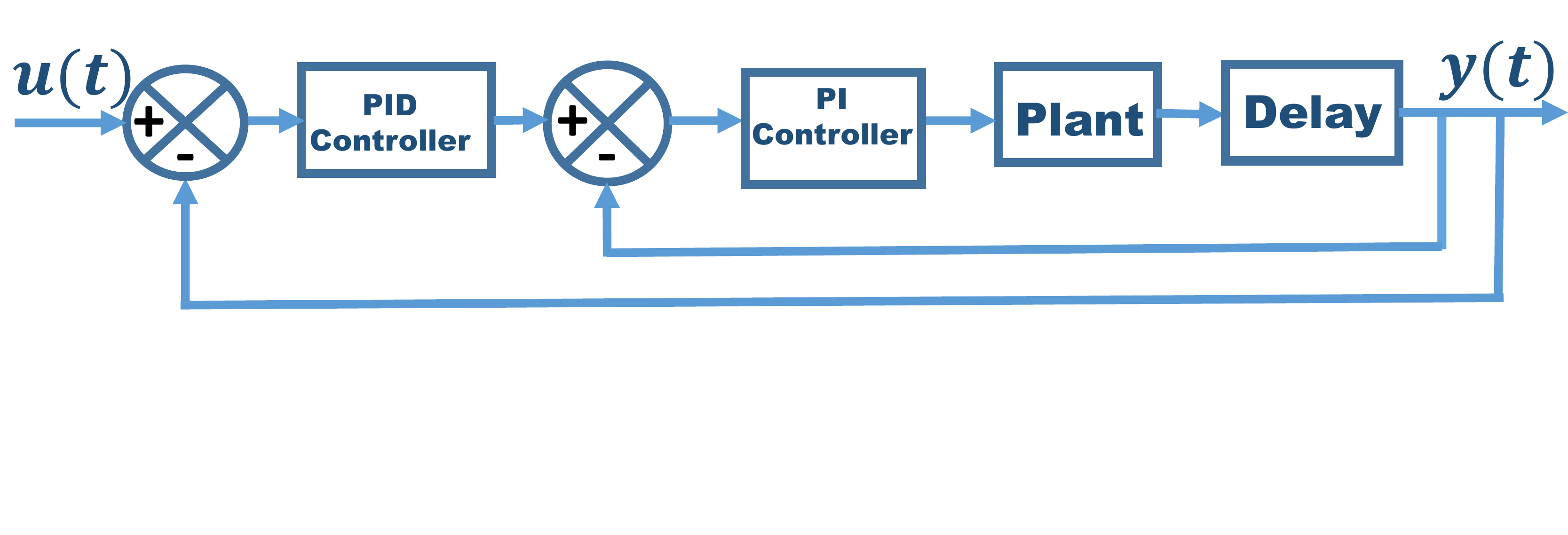}	
		\vspace{-4.6em}
		\caption{Block Diagram of Model}
		\vspace{-1.1em}
		\label{Fig. Model.}
	\end{figure} 
	\begin{figure}[tb]
		\centering
		\includegraphics[trim=30mm 0mm 30mm 0mm, width=3.4in, clip=true]{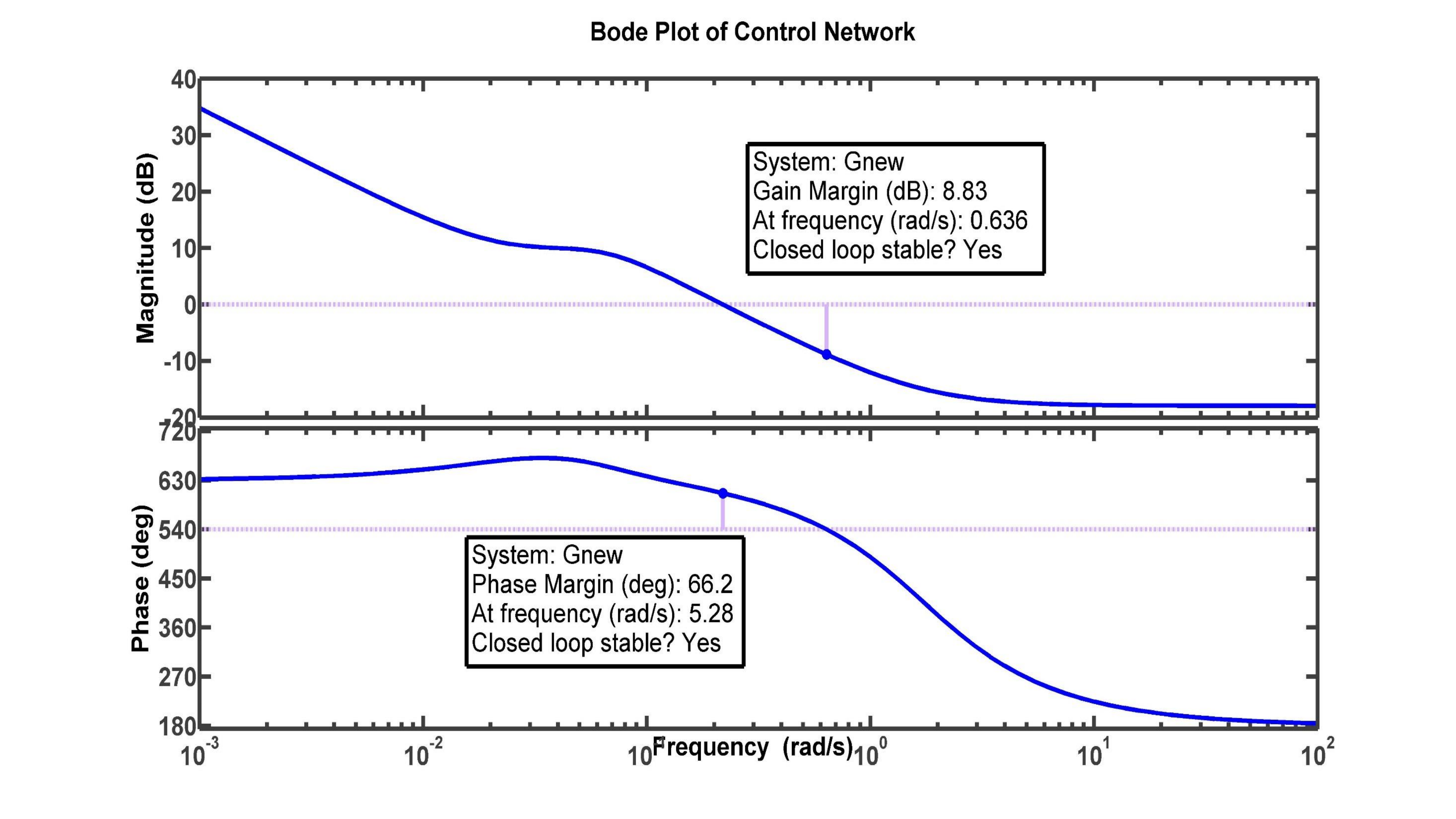}
		\caption{Bode Response Plot of feedforward and Cascade Control Network}
		\label{Fig. 21.}
	\end{figure}  
	  
	\section{Experimental Validation} \label{sec. expt}
	The testbed of Fig. \ref{Fig. 1.} was used to validate the proposed model and control network. The control algorithm was implemented using a Runge-Kutta second-order ordinary differential equation solver that ran on the myRIO at a fixed step size of 0.1 second. A fixed step solver was used to avoid reduction in computational efficiency in having to discretize the controller and soft robot model, which were both modeled in the continuous-time domain. To ensure the deployment was executed in real-time, the timing source of the execution loop on the Windows workstation was also synchronized to the myRIO hardware.
		\begin{figure}[tb]
		\centering
		\includegraphics[keepaspectratio = true, width=3.6in, clip=true]{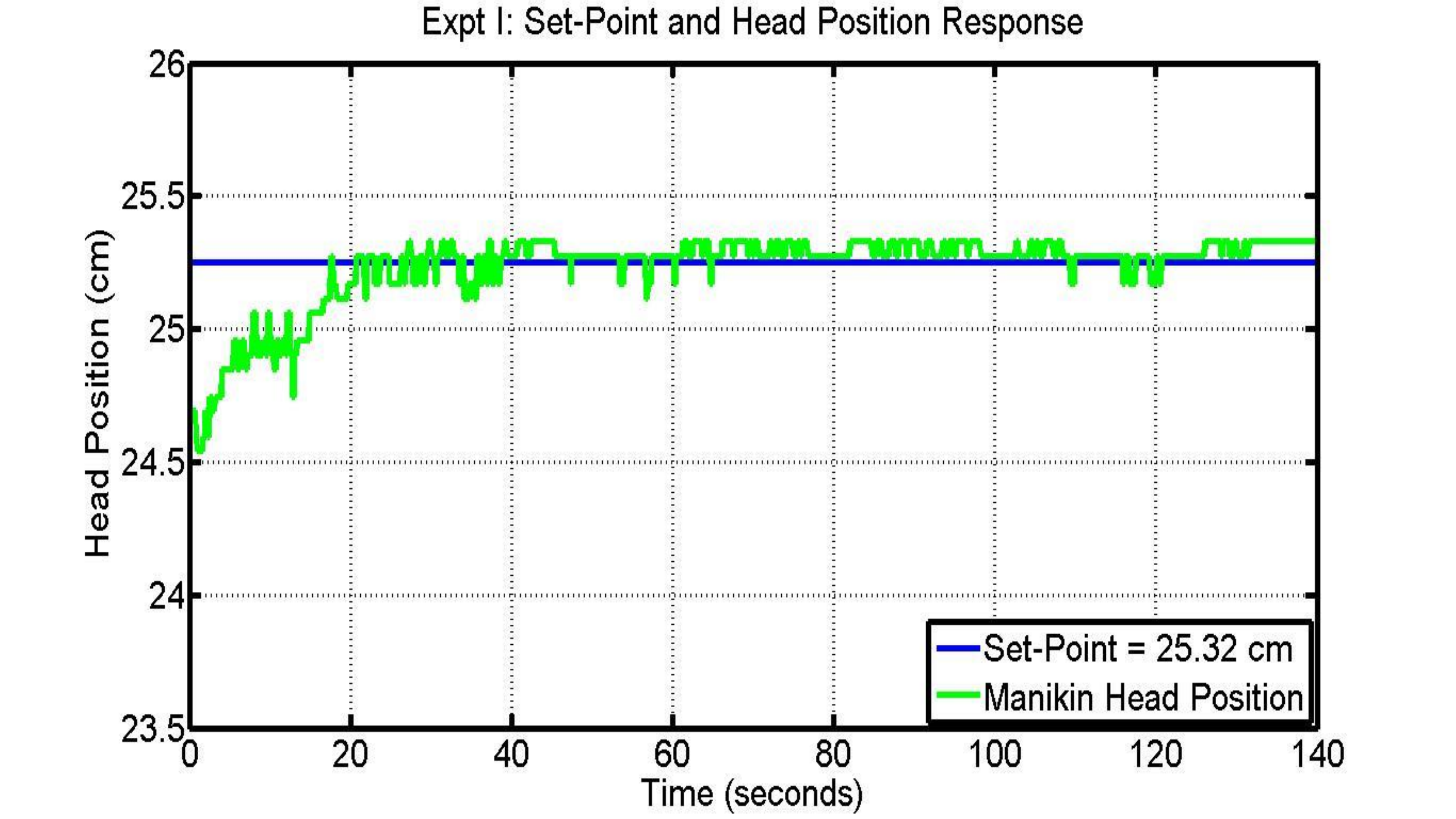}
		\vspace{0.1cm}
		\caption{Manikin head response to a constant setpoint}
		\label{Fig. 23a.}
	\end{figure}
	\begin{figure}[tb]
		\centering
		\includegraphics[keepaspectratio = true, width=3.6in, clip=true]{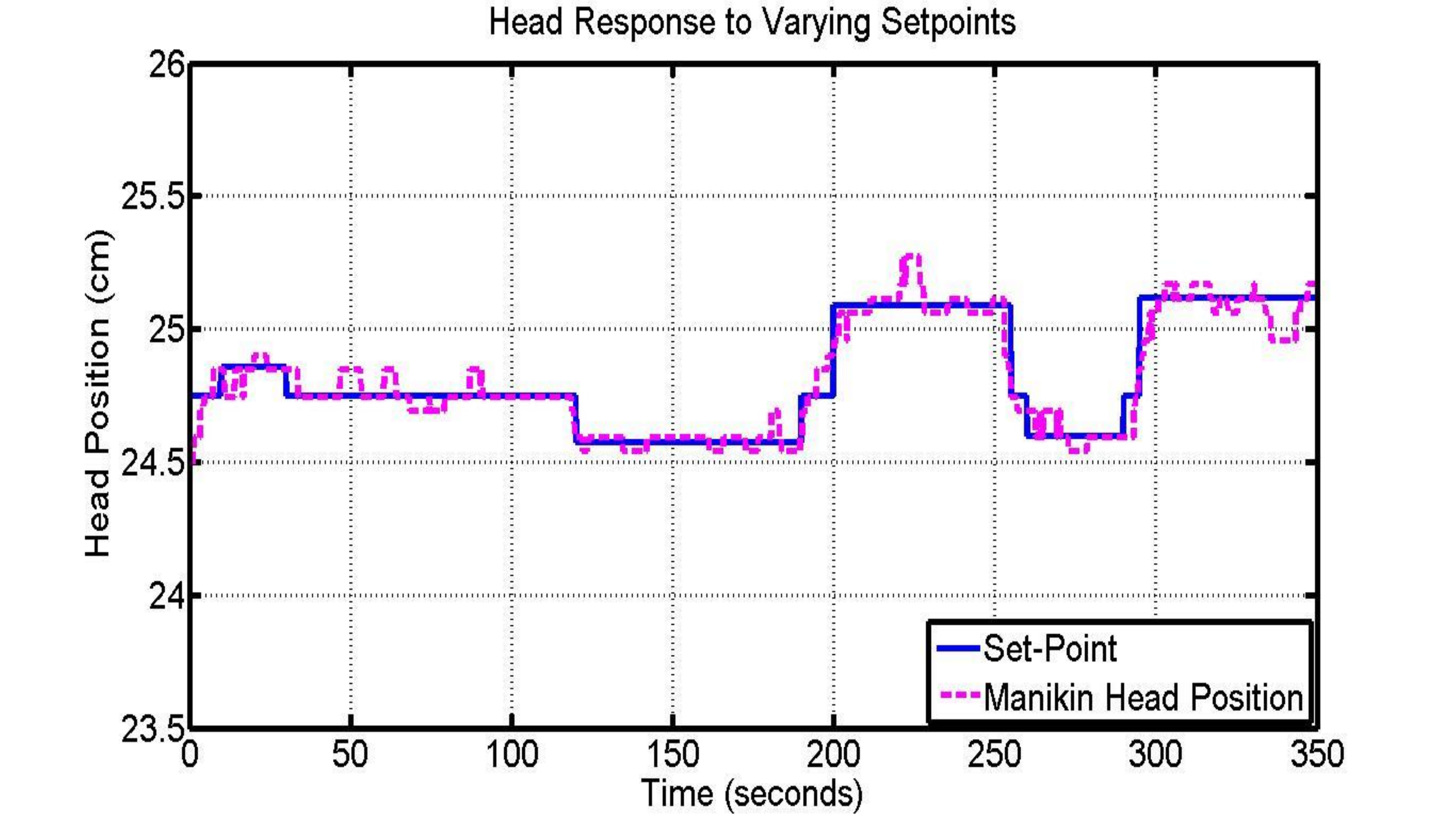}
		\caption{Varying Set-points and Manikin Head Trajectory Tracking}
		\label{Fig. 23b.}
	\end{figure}

	Experimental results with constant reference tracking is shown in Fig. \eqref{Fig. 23a.}. With a constant set-point target of 25.32cm above the table, and the manikin head being 24.51cm above the table at rest position, the algorithm was deployed to track the set-point trajectory. The controller behaves as expected, reaching within 2\% of the reference after a rise time of approximately 15 seconds and tracking the setpoint trajectory to within 0.2cm maximum deviation. The system also displays less overshoot and clinically acceptable settling time. A second experiment (Fig. \ref{Fig. 23b.})with changing set-points was carried out. The controller tracks the set-point trajectories with a maximum deviation of 2mm from setpoint. 
	
	The depth and range resolution of the Kinect Xbox sensor accounts for the deviation from setpoint trajectory when the controller is applied. The chosen set-points of figures \eqref{Fig. 23a.}, and \eqref{Fig. 23b.} are extensible for use in target clinical applications as a typical H\&N RT may demand. Future multi-axis head positioning work will explore the new time-of-flight based Kinect for Windows v2 sensor which has an improved noise floor, visualizes small objects in greater detail and more clearly, and a  depth fidelity of 512 $\times$ 424 pixels and a wider field of view (fov) of 70.6 $\times$ 60 degrees compared with the  320 $\times$ 240 pixels with 58.5 $\times$ 46.6 degrees fov of the Xbox sensor used in this work.

	\section{CONCLUSIONS}  \label{conclusions}
	Accurate positioning of the patient head and torso is crucial in intensity modulated radiotherapy. Deviations from desired positions have been known to cause dose variation, degenerate treatment efficacy, brain necrosis and edema\cite{c7}. In this paper, the control of cranial flexion/extension motion of a patient during maskless and frameless, image-guided radiotherapy was considered using a manikin head as a test subject. We established that the proposed soft robot can track a desired step reference trajectory with 2mm precision after a lag time of 15 seconds. This was achieved using a PI controller nested within a PID feedforward configuration and implemented on an NI myRIO. The Kinect Xbox 360 sensor sensed head position. 
	
	This shows the possibility of accurate positioning with the choice of a deliberate, well-tuned controller. Future efforts will focus on designing a more accurate and robust controller usable for clinical RT and improve the transient characteristics.  We will also look into gain scheduling to allow different settling times for different motion, as fast motions may be uncomfortable for patients. Long term efforts include extending the results to the deformable motions of the upper torso, and H\&N. This would involve multiple bladders, finding the coupling needed between IABs to give desired actuation, refining the system model for the bladders, and developing a more accurate and robust controller, in order to achieve multi-axis positioning irrespective of patient head shape or size. This would demonstrate comprehensive and accurate  automated control of a patient's position during cancer H\&N radiotherapy, prevent unwanted anatomical deformations and other harmful effects that positioning deviations have been known to cause.
	
	\addtolength{\textheight}{-12cm}   
	


	


\end{document}